\documentclass[a4paper]{article}

\usepackage[english]{babel}
\usepackage[utf8x]{inputenc}
\usepackage[T1]{fontenc}
\usepackage{multicol}
\usepackage[a4paper,top=3cm,bottom=2cm,left=3cm,right=3cm,marginparwidth=1.75cm]{geometry}
\usepackage{mlhc}

\usepackage{times}
\usepackage{graphicx} 
\usepackage{natbib}
\usepackage{algorithm}
\usepackage{algorithmic}
\usepackage{url}
\usepackage{amsmath}
\usepackage{amsmath}
\usepackage{amsfonts}
\usepackage{amssymb}
\usepackage{graphicx}
\usepackage[colorinlistoftodos]{todonotes}
\usepackage{glossaries}
\usepackage{nicefrac}
\usepackage{bbm}
\usepackage{appendix}
\usepackage{booktabs}
\usepackage{subcaption}
\usepackage{wrapfig}

\usepackage[colorinlistoftodos]{todonotes}

\title{Predicting Surgery Duration with \\ 
Neural Heteroscedastic Regression}

\author{Nathan H Ng$^1$, Rodney A Gabriel$^{2,3}$, Julian McAuley$^1$, Charles Elkan$^1$, Zachary C Lipton$^{1,2}$ \thanks{Corresponding author, website: http://zacklipton.com} \\
Department of Computer Science$^1$ \\ 
Division of Biomedical Informatics$^2$\\
Department of Anesthesiology$^3$ \\ 
University of California, San Diego\\
9500 Gilman Drive
La Jolla, CA 92093, USA \\
\texttt{\{nhng, ragabriel, jmcauley, elkan, zlipton\}@ucsd.edu}
}
\date{}
\begin{document}

\maketitle

\begin{abstract}
Scheduling surgeries is a challenging task 
due to the fundamental uncertainty of the clinical environment, 
as well as the risks and costs associated with under- and over-booking.
We investigate neural regression algorithms 
to estimate the parameters of surgery case durations, 
focusing on the issue of \emph{heteroscedasticity}. 
We seek to simultaneously estimate the duration of each surgery, 
as well as a surgery-specific notion 
of our \emph{uncertainty} about its duration.  
Estimating this uncertainty can lead 
to more nuanced and effective scheduling strategies, 
as we are able to schedule surgeries more efficiently 
while allowing an informed and case-specific margin of error. 
Using surgery records 
from a large United States health system
we demonstrate potential improvements on the order of 
20\% (in terms of minutes overbooked) compared to current scheduling techniques.
Moreover, we demonstrate that surgery durations are indeed heteroscedastic.
We show that models that estimate case-specific uncertainty better fit the data (log likelihood). 
Additionally, we show that the heteroscedastic predictions can more optimally trade off between over and under-booking minutes, especially when idle minutes and scheduling collisions confer disparate costs.
\end{abstract}

\section{Introduction}
\label{sec:introduction}
In the United States, 
healthcare is expensive and hospital resources are scarce.  
Healthcare expenditure now exceeds $17\%$ of US GDP \citep{expenditure2012},
even as surgery wait times appear to have increased over the last decade \citep{bilimoria2011wait}.
One source of inefficiency (among many)
is the inability to fully utilize hospital resources.
Because doctors cannot 
accurately predict 
the duration of surgeries, 
operating rooms can become congested (when surgeries run long) or lie vacant (when they run short).
Over-booking can lead to long wait times and
higher costs of labor (due to over-time pay),
while under-booking decreases throughput,
increasing the marginal cost per surgery. 

At present, doctors book rooms according to a simple formula. The default time reserved is simply the mean duration of that specific procedure. 
The procedure code does in fact explain 
a significant amount of the variance 
in surgery durations.
But by ignoring other signals, we hypothesize 
that the medical system leaves important signals untapped. 
 
We address this issue by developing better and more nuanced strategies for surgery case prediction.
Our work focuses on a collection of surgery logs
recorded in Electronic Health Records (EHRs)
at a large United States health system.
For each patient, we consider a collection of pre-operative features, including patient attributes 
(age, weight, height, sex, co-morbidities, etc.),
as well as attributes of the clinical environment, 
such as the surgeon, surgery location, and time.
For each procedure, we also know how much time was originally scheduled, in addition to the actual \emph{`ground-truth'} surgery duration, 
recorded after each procedure is performed.

\begin{figure*}[t]    
  		\centering
		\includegraphics[scale=.5]{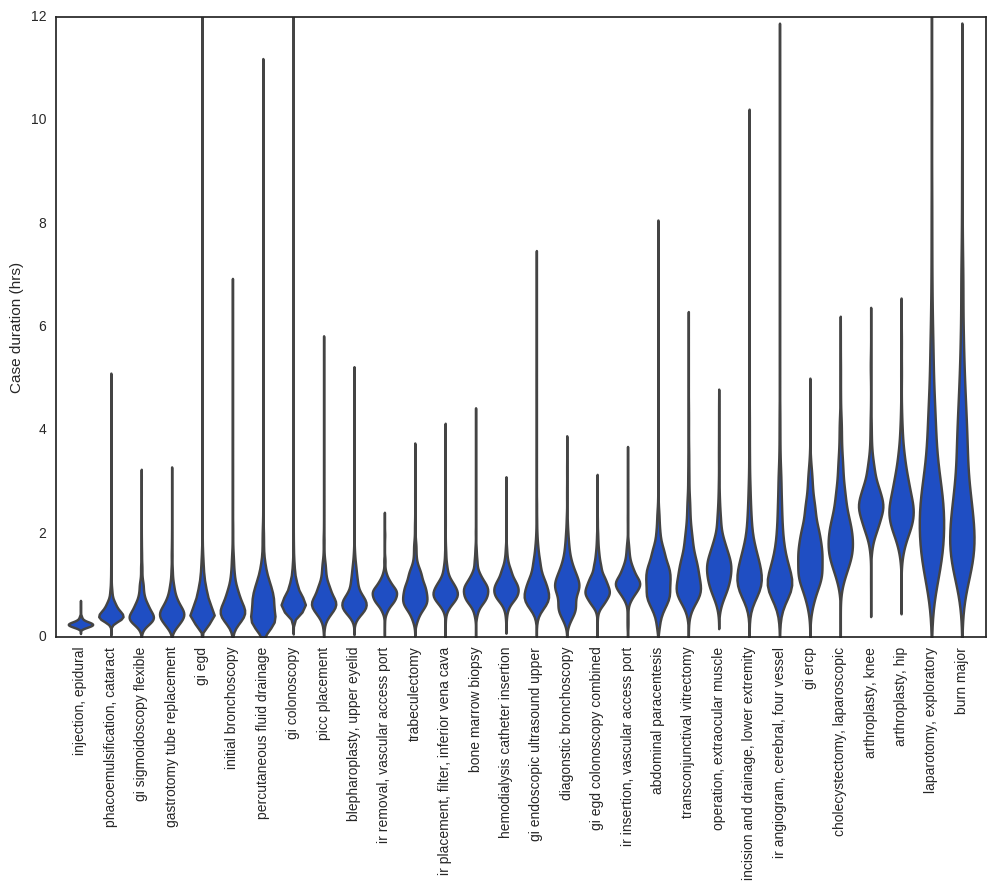}
		\caption{Distributions of durations 
        for the $30$ most common procedures.}
\label{fig:surgery-duration-violin}
\end{figure*}

We are particularly interested in developing methods that allow us to better estimate the \emph{uncertainty} associated with the duration of each surgery.
Typically, neural network regression objectives assume \emph{homoscedasticity}, i.e., 
constant levels of target variability for all instances.
While mathematically convenient, 
this assumption is clearly violated in data such as ours:
as one might surmise intuitively that
operations that typically take a long time
tend to exhibit greater variance
than shorter ones.
For example, among the $30$ most common procedures, 
\emph{epidural injections} are both the shortest procedures and the ones with the least variance (Figure \ref{fig:surgery-duration-violin}). 
Among the same $30$ procedures, \emph{exploratory laparotomy} and \emph{major burn surgery} exihibit the greatest variance. All procedures exhibit long (and one-sided) tails. 

To model this data, we revisit the idea 
of heteroscedastic neural regression, 
combining it with expressive, 
dropout-regularized neural networks.
In our approach, we jointly learn all parameters 
of a predictive distribution. 
In particular, we consider Gaussian and Laplace distributions, each of which is parameterized 
by a mean and standard deviation. 
We also consider Gamma distributions, which are especially suited to survival analysis.
Unlike the Gaussian and Laplace which are long tailed on both ends, the gamma has a long right tail and 
has only positive support (i.e., it
assigns zero probability density 
to any value less than zero).
The restriction to positive values suits the modeling of durations or other survival-type data. 
While the gamma distribution (and the related Weibull distribution) has been applied to medical data with classical approaches \cite{bennett1983log,sahu1997weibull},
this is, to our knowledge, the first to approximate the a parameters of a gamma
distribution using modern neural network approaches. 

Our heteroscedastic models better fit the data (as determined by log likelihood) compared to both
current practice and neural network baselines that fail to account for heteroscedasticity.
Furthermore, our models produce
reliable estimates of the variance, 
which can be used to schedule intelligently, especially when over-booking and under-booking confer disparate costs. 
These uncertainty estimates 
come at no cost in performance by traditional measures. 
The best-performing Gamma MLP model 
achieves a lower mean squared error 
than a vanilla least squares (Gaussian) MLP, 
despite optimizing a different objective. 

\section{Dataset}
\label{sec:dataset}
Our dataset consists of patient records extracted from the EHR system at a large 
United States
hospital.
Specifically, we selected 107,755 records
corresponding to surgeries that took place
between 2014 and 2016.
These surgeries span 995 distinct procedures,
and were performed by 368 distinct surgeons. 
Histograms of both are long-tailed, 
with over $796$ procedures performed 
fewer
than $100$ times 
and $213$ doctors performing 
fewer
than $100$ surgeries each.
Moreover the data contains several clerical mistakes in logging the durations.
For example, a number of surgeries in the record were reported as running less than $5$ minutes.
Discussions with the hospital experts suggest 
that this may indicate either clerical errors 
or an inconsistently applied convention for logging canceled surgeries.
Additionally, several surgeries were reported to run over $24$ hours, suggesting (rare) clerical errors in logging the end times of procedures. 
We remove all surgeries reported to take less than $5$ minutes or more than $24$ hours from the dataset.
This preprocessing left us with roughly 80\% of our original data (86,796 examples).
For our experiments, we split this remaining data 80\%/8\%/12\% for training/validation/testing.

\subsection{Inputs}
For each surgery, we extracted a number of pre-operative features from the corresponding EHRs. 
We restrict attention to features that are available 
for a majority of patients and (to avoid target leaks) exclude any information 
that is charted during or following the procedure.
Our features fall into several categories: 
patient, doctor, procedure, and context. 

\paragraph{Patient features:}
For each of our patients, we include the following features:
\begin{itemize}
\item \textbf{Size:} Patient height and weight are real-valued features. We normalize each to mean $0$, variance $1$. 
\item \textbf{Age:} A categorical variable,
binned according to ten-year wide intervals that are open on the left side $(0-10], (10-20],\ldots$
None of the patients in our cohort are zero years old.
\item \textbf{ASA score:} an ordinal score that represents the severity of a patient's illness. For example, ASA I denotes a healthy patient, ASA III denotes severe systemic disease, and ASA V denotes that the patient is moribund without surgery. ASA VI refers to a brain-dead patient in preparation for organ transplantation.
\item \textbf{Anesthesia Type: } This categorical feature represents the type of anesthesia applied to sedate the patient. The values assigned to this variable include \emph{General}, \emph{Monitored anesthesia care (MAC)}---in which a patient undergoes local anesthesia together with sedation, \emph{Neuraxial}, \emph{No Anesthesiologist}, and \emph{other/unknown}.
\item \textbf{Patient Class: } This categorical feature indicates the patient's current status. The values assigned to this variable include \emph{Emergency Department Encounter}, \emph{Hospital Outpatient Procedure}, \emph{Hospital Outpatient Surgery}, \emph{Hospital Inpatient Surgery}, \emph{Trauma Inpatient Admission}, \emph{Inpatient Admission}, \emph{Trauma Outpatient}. 
\item \textbf{Comorbidities:} We model the following co-morbidities as binary variables: smoker status, atrial fibrillation, chronic kidney disease, chronic obstructive pulmonary disease, congestive heart failure, coronary artery disease, diabetes, hypertension, cirrhosis, obstructive sleep apnea, cardiac device, dialysis, asthma, and dementia.
\end{itemize}

\paragraph{Doctor:} 
We represent the doctor performing the procedure (categorical) using a one-hot vector. 
The \emph{doctor} feature exhibits considerable class imbalance,
with the most prolific doctor performing $3770$ surgeries
and the least prolific doctor (in the pruned dataset) performing $100$.

\paragraph{Procedure: } 
We represent the procedure performed as a one-hot vector. The most common operations tend to be minor GI procedures: the four most frequent procedures are 
\emph{colonoscopy}, 
\emph{upper GI endoscopy}, 
\emph{cataract removal}, 
and \emph{abdominal paracentesis}.
This distribution is also long-tailed with 11,173 colonoscopies.

\paragraph{Context:} 
We represent the context of the procedure with several categorical variables. First, we represent the hour of the day as a categorical variable with values binned into $8$ non-overlapping $3$-hour width buckets. 
Second, we represent the day of the week and month of the year each as one-hot vectors. 
Finally, we similarly represent the location of the operations as a one-hot vector.\\

We summarize the number and kind of features in our dataset in Table \ref{tab:dataset}.
We handle variables with missing values, including height, weight, and hour of the day, 
by incorporating missing value indicators, 
following previous work on clinical datasets \citep{lipton2016modeling}.

\begin{table*}[t]
\begin{centering}
\begin{tabular}{lllll}
\toprule
\textbf{\textbf{Feature}} & \textbf{Abbreviation} & \textbf{Type} & \textbf{\#Categories} & \textbf{Mode}  \\
\midrule
\textbf{Age} & age & Categorical &$9$ &$50$-$60$ \\
\textbf{Sex} & sex & Binary & $2$ & Female\\
\textbf{Weight} & weight & Numerical & -&- \\
\textbf{Height} & height & Numerical & -&- \\
\textbf{Time of Day} & hour & Categorical & $8$ & 9:00-12:00  \\
\textbf{Day of Week} & day & Categorical & $7$ & Friday \\
\textbf{Month} & month & Categorical & $12$ & March\\
\textbf{Location} & location & Categorical & $10$ & 
- \\
\textbf{Patient Class} & class & Categorical & $7$& Hospital Outpatient\\
\textbf{ASA Rating} & asa & Categorical & $6$ & None\\
\textbf{Anesthesia Type} & anesthesia & Categorical & $5$ & General \\
\textbf{Surgeon} & surgeon & Categorical & 155 & - \\
\textbf{Procedure} & procedure & Categorical & 199 & Colonoscopy \\
\textbf{Smoker} & smoker & Binary & $2$ & No\\
\textbf{Heart Arrhytmia} & afib & Binary & $2$ & No \\
\textbf{Chronic Kidney Disease} & ckd & Binary & $2$ & No \\
\textbf{Congestive Heart Failure} & chf & Binary & $2$ & No\\
\textbf{Coronoary Artery Disease} & cad & Binary & $2$ & No\\
\textbf{Type II Diabetes} & diabetes & Binary & $2$ & No\\
\textbf{Hypertension} & htn & Binary & $2$ & No\\
\textbf{Liver Cirrhosis} & cirrhosis & Binary & $2$ & No\\
\textbf{Sleep Apena} & osa & Binary & $2$ & No\\
\textbf{Cardiac Device} & cardiac\_device & Binary & $2$ & No\\
\textbf{Dialysis} & dialysis & Binary & $2$ & No\\
\textbf{Asthma} & asthma & Binary & $2$ & No\\
\textbf{Dementia} & dementia & Binary & $2$ & No\\
\textbf{Cognitive Impairment} & cognitive & Binary & $2$ & No\\
\bottomrule
\end{tabular} 
\vspace{-2mm}
\caption{
Summary of features.}
\label{tab:dataset}
\end{centering}
\vspace{-2mm}
\end{table*}

\section{Methods}
\label{sec:methods}
This paper addresses the familiar task of regression.
We start off by refreshing some basic preliminaries.
Given a set of examples $\{\boldsymbol{x}_i\}$, and corresponding labels $\{y_i\}$, we desire a model $f$ that outputs a prediction $\hat{y} = f(\boldsymbol{x})$. 
The task of the machine learning algorithm is to produce the function $f$ given a dataset $\mathcal{D}$ consisting of examples $X$ and labels $\boldsymbol{y}$.
Generally, we seek predictions that are somehow \emph{close} to $y$, as determined by some computable \emph{loss function} $\mathcal{L}$. 
Most often we minimize the squared loss $\mathcal{L} = \sum_{i} (y_i - \hat{y}_i)^2$ for all instances ($\boldsymbol{x}_i$, $y_i$). 

One popular method for producing such a function 
is to choose a class of functions $f$ 
parameterized by some values $\boldsymbol{\theta}$. 
Linear models are the simplest examples of this approach.
To train a linear regression model, we define $f(\boldsymbol{x})=\theta^T\boldsymbol{x}$.
Then we solve the following optimization problem:
$$\boldsymbol{\theta}^* = 
\text{argmin}_{\boldsymbol{\theta}} 
\mathcal{L}(\boldsymbol{y}, \hat{\boldsymbol{y}})
$$
over some training data and evaluate the model by its performance on previously unseen data.
For linear models, the error-minimizing parameters (on the training data) can be calculated analytically. 
For all modern deep learning models, no analytic solution exists, so optimization
typically
proceeds by stochastic gradient descent.

For neural network models, we change only the function $f$.
In multilayer perceptrons (MLP) for example, we transform our input through a series of matrix multiplications, 
each followed by a nonlinear activation function.
Formally, an $L$-layer MLP for regression has the simple form
$$\boldsymbol{\hat{y}}= W_L \cdot \phi(W_{L-1} 
\cdot \ldots \cdot
\ \phi(W_1 \cdot \boldsymbol{x} + b_1) + \ldots + b_{L-1}) + b_L\,,$$ 
where $\phi$ is an activation function such as sigmoid, tanh, or rectified linear unit (ReLU) and $\boldsymbol{\theta}$ consists of the full set of parameters $W_l$ and $b_l$. 

We might view the loss function (squared loss) 
as simply an intuitive measure of distance. 
Alternatively, it's possible to derive the choice of squared loss by viewing regression from a probabilistic perspective. 
In the probabilistic view, a parametric model outputs a distribution $P(y|\boldsymbol{x})$.

In the simplest case, we can assume that the prediction $\boldsymbol{\hat{y}}$ is the mean of a Gaussian predictive distribution with some variance $\sigma$.
In this view, we can calculate the probability density of any $y$ given $\boldsymbol{x}$, and thus can choose our parameters according to the maximum likelihood principle:
\begin{equation}
\boldsymbol{\theta}^{\text{MLE}}
=\max_{\theta} \prod_{i=1}^{n} \frac{1}{\sqrt{2\pi \hat{\sigma}^2}} \exp \left( \frac{-(y_i-\hat{y}_i)^2}{2\hat{\sigma}^2} \right)
= \min_{\boldsymbol{\theta}} \sum_{i=1}^{n} 
\left( \log(\hat{\sigma}_i) + \frac{(y_i-\hat{y}_i)^2}{2\hat{\sigma}^2}  \right).
\end{equation}
Assuming constant $\hat{\sigma}$, 
this yields a familiar least-squares 
objective.

In this work, we relax the assumption of constant variance (homoscedasticity), 
predicting both $\hat{y}(\theta, \boldsymbol{x})$ and $\hat{\sigma}(\theta, \boldsymbol{x})$ simultaneously.
While we apply the idea to MLPs,
it is easily applied to networks of arbitrary architecture. 
To predict the standard deviation $\hat{\sigma}$ of the predictive distribution, we modify our MLP to have two outputs:
The first output has linear activation and we interpret its output as the conditional mean $\hat{y}$.
The second output models the conditional variance $\hat{\sigma}$.  
To enforce positivity of $\hat{\sigma}$, 
we run this output through the softplus activation function $\text{softplus}(z) = \log(1 + \exp(z))$. 

We extend the same idea to Laplace distributions, 
which turn out to better describe 
the target variability for surgery duration, 
and are also maximum likelihood estimators when optimizing the Mean Absolute Error (MAE).
Mean Absolute Error corresponds to the average number of minutes over or underbooked, 
and is typically the quantity of interest for this type of scheduling task.
The Laplace distribution is parameterized by $b=\sqrt{2}\sigma$:
\begin{equation}
\boldsymbol{\theta}^{\text{MLE}}
= \max  \prod_{i=1}^{n} 
\frac{1}{2b} 
\exp \left( \frac{- |y_i - \hat{y}_i|}{b} \right)
=\min_{\theta} \sum_{i=1}^n \left( \log{b} + \frac{|y_i - \hat{y}_i|}{b} \right).
\end{equation}

Finally, we apply the same technique to perform neural regression with gamma predictive distributions. 
The gamma distribution
has strictly positive support and is long-tailed on the right.
Since surgeries and other survival-type data 
have nonnegative lengths, 
probability distributions 
with similarly nonnegative support such as the gamma distribution (compared to the real-valued support of the Gaussian and Laplace distributions), 
might better describe surgery duration.
Formally, the expected time between surgeries (or their associated durations) follows a gamma distribution when surgery start times are modeled as a Poisson process.

The gamma distribution is parametrized by a shape parameter $k$ and a scale parameter $\Phi$:
\begin{equation}
\boldsymbol{\theta}^{\text{MLE}}
= \max \prod_{i=1}^{n} \frac{1}{\Gamma(k)\Phi^k}
y_i^{k-1}\exp \left( \frac{-y_i}{\Phi} \right) 
=\min_{\theta} \sum_{i=1}^n \left( \log(\Gamma(k)) + k\log\Phi - (k-1) \log y_i + \frac{y_i}{\Phi} \right).
\end{equation}
In this case, the model now needs to predict two values: $\hat{k}(\theta, x)$ and $\hat{\Phi}(\theta, x)$. 
As before, our MLP has two outputs, 
with both passed through a softplus activation 
to enforce positivity.

\section{Experiments}
\label{sec:experiments}
We now present the basic experimental setup. 
For all experiments we use the same $80\%$/$8\%$/$12\%$ training/validation/test set split. 
Model weights are updated on the training set 
and we choose all non-differentiable hyper-parameters 
and architecture details 
based on validation set performance.
In the final tally, we have $441$ features, the majority of which are sparse and accounted for by the one-hot representations over procedures and doctors.
We express our labels (the surgery durations)
as the number of hours that each procedure takes.

\paragraph{Baselines}
We consider three sensible baselines for comparison.
The first is to follow the current heuristic 
of predicting the average time per procedure.
Note that this is equivalent to training an unregularized linear regression model with a single feature per procedure and no others.
Although the main technical contribution of this paper
is concerned with
modeling heteroscedasticity, 
we are also generally interested to know how much performance the current approach leaves untapped. This baseline helps us to address this question. 
We also compare against linear regression. 
While we tried applying $\ell_2$ regularization,
choosing the strength of regularization $\lambda$ on holdout data, this did not lead to improved performance. 
Finally, we compare against traditional multilayer perceptrons.
To calculate NLL for models that assume homoscedasticity, we choose the constant variance that minimizes NLL on the validation set. 

\paragraph{Training Details}
For all neural network experiments, 
we use MLPs with ReLU activations.
We optimize each network's parameters 
by stochastic gradient descent,
halving the learning rate every $50$ epochs. 
For each experiment, we used an initial learning rate of $.1$.
To determine the architecture, we performed a grid search over the number of hidden layers (in the range 1-3) and over the number of hidden nodes, 
choosing between {128, 256, 384, 512}.
As determined by our hyper-parameter optimization, 
for homoscedastic models, all MLPs use $1$ hidden layer with $128$ nodes.
All heteroscedastic models use $1$ hidden layer with $256$ nodes.
All models use dropout regularization.

For our basic quantitative evaluation, 
we report the root mean squared error (RMSE), 
mean absolute error (MAE), 
and negative log-likelihood (NLL). 
For heteroscedastic models, 
we evaluate NLL using the predicted parameters of the distribution.
For the Gamma distribution, we calculate its mean as $k \cdot \Phi$. We use its mean as the prediction $\hat{y}$ for calculating RMSE. To calculate MAE, we use the median of the Gamma 
distribution
as $\hat{y}$. Although the median of a Gamma distribution has no closed form, 
it can be efficiently approximated.

We summarize the results in Table \ref{tab:prediction-results}. The heteroscedastic Gamma MLP performs best 
as measured 
by both RMSE and NLL, while the Laplace MLP performs 
best
as measured by MAE.
All heteroscedastic models outperform all homoscedastic models (as determined by NLL) with the heteroscedastic Gamma MLP achieving an NLL of $.4668$ as compared to $1.062$ by the best performing homoscedastic model (Laplace MLP). The significant quantity when evaluating log likelihood is the \emph{difference} between NLL values, corresponding to the (log of the) likelihood ratio between two models.

\begin{figure*}[t!]
	\centering   
\begin{subfigure}[b]{0.3\textwidth} 
  		\centering
		\includegraphics[width=\linewidth]{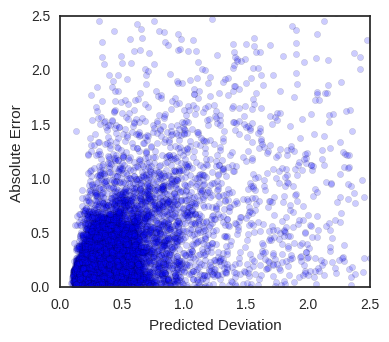}
		\caption{Gaussian}
\end{subfigure}
 \begin{subfigure}[b]{0.3\textwidth}
  		\centering
		\includegraphics[width=\linewidth]{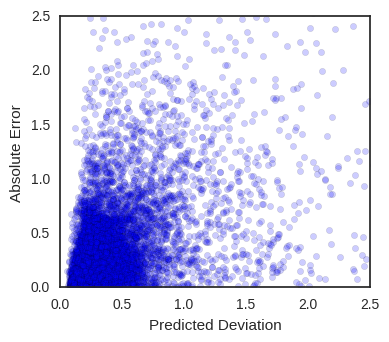}
        \caption{Laplacian}
\end{subfigure}
\begin{subfigure}[b]{0.3\textwidth}
  		\centering
		\includegraphics[width=\linewidth]{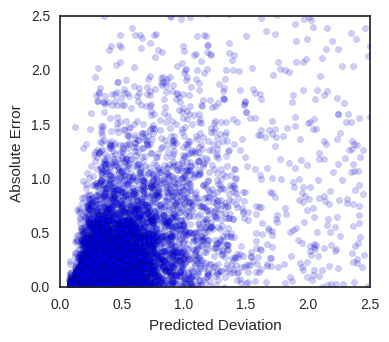}
        \caption{Gamma}
\end{subfigure}
\begin{subfigure}[b]{0.3\textwidth} 
  		\centering
		\includegraphics[width=\linewidth]{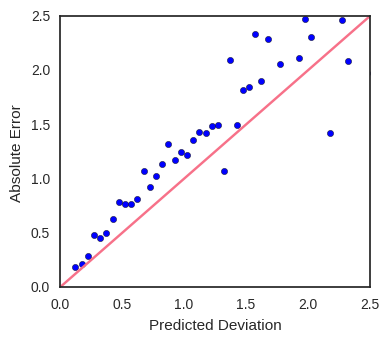}
		\caption{Gaussian}
\end{subfigure}
\begin{subfigure}[b]{0.3\textwidth} 
  		\centering
		\includegraphics[width=\linewidth]{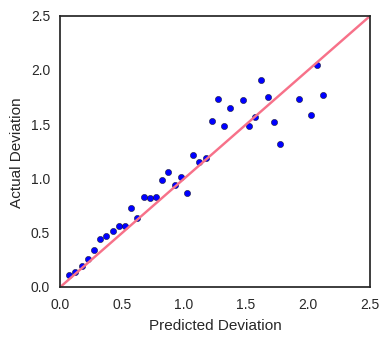}
		\caption{Laplacian}
\end{subfigure}
\begin{subfigure}[b]{0.3\textwidth} 
  		\centering
		\includegraphics[width=\linewidth]{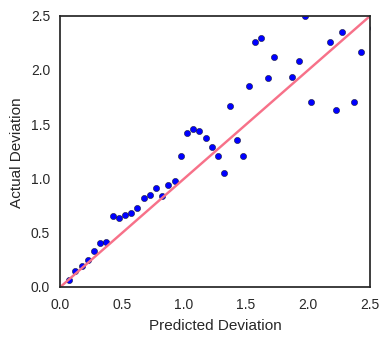}
		\caption{Gamma}
\end{subfigure}
 \caption{Plots of predicted $\hat{\sigma}$ against absolute error with heteroscedastic Gaussian (a), Laplacian (b), and Gamma (c) models. Averaging over bins of width $0.05$ (d) (e) (f), shows that $\hat{\sigma}_i$ is a reliable estimator of the observed error.  
}
\label{fig:deviation-error}
\end{figure*}

\begin{figure*}[t!]
	\centering
    
\begin{subfigure}[b]{0.5\textwidth} 
  		\centering
		\includegraphics[width=0.7\linewidth]{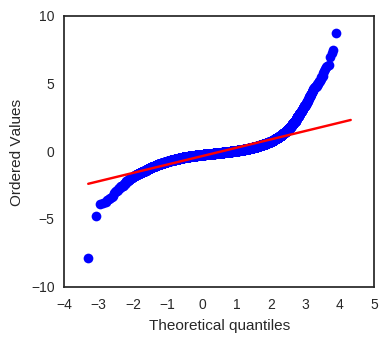}
		\caption{Gaussian QQ Plot}
\end{subfigure}~
    \begin{subfigure}[b]{0.5\textwidth}
  		\centering
		\includegraphics[width=0.7\linewidth]{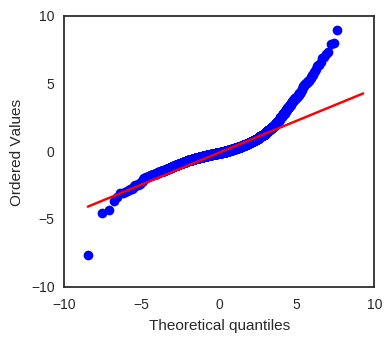}
        \caption{Laplace QQ Plot}
\end{subfigure}
 \caption{QQ plots of observed error for Gaussian and Laplace noise models. The Laplace distribution better describes observed error, with shorter tails at both ends.
}
\label{fig:qq}
\end{figure*}

Plots in Figure \ref{fig:deviation-error} demonstrate that the predicted deviation reliably estimates the observed error, and QQ plots (Figure \ref{fig:qq}) demonstrate that the Laplace distribution appears to fit our targets better than a Gaussian predictive distribution. This gives some (limited) insight into why the Laplace predictive distribution might better fit our data than the Gaussian. 

\begin{table}
\begin{centering}
\begin{tabular}{lcccc}
\toprule
\textbf{\textbf{Models}} & \textbf{RMSE} & \textbf{MAE} & \textbf{NLL} & \parbox{0.2\linewidth}{\centering \textbf{Change in NLL vs.~Current Method}} \\
\midrule
\textbf{Current Method} & $49.80$ &$28.87$ &$1.2385$ & $0.0000$ \\
\textbf{Procedure Means}   &  $49.06$ &$27.70$ &$1.2222$ & $0.0164$
  \\
\textbf{Linear Regression} & $45.23$ &$25.07$ &$1.1446$ & $0.0939$\\
\textbf{MLP Gaussian } & $43.51$ &$23.90$ &$1.1102$ & $0.1283$\\
\textbf{MLP Gaussian HS} & $44.03$ &$24.23$ &$0.7325$ & $0.5060$\\
\textbf{MLP Laplace } & $44.24$ &$\boldsymbol{23.14}$ &$1.0621$ & $0.1765$\\
\textbf{MLP Laplace HS} & $45.07$ &$23.41$ &$0.5034$ & $0.7351$\\
\textbf{MLP Gamma HS} &$\boldsymbol{43.38}$ &$23.23$ &$\boldsymbol{0.4668}$ & $0.7717$\\
\bottomrule
\end{tabular} 
\vspace{-2mm}
\caption{
Performance on test-set data (lower is better). 
MLP models outperform alternatives at the 1\% significance level or better.
}
\label{tab:prediction-results}
\end{centering}
\end{table}

\section{Related Work}
\label{sec:related}
Previous work in in the medical literature addresses the prediction of surgery duration
\citep{eijkemans2010predicting, kayics2015robust, devi2012prediction},
accounting for both patient and surgical team characteristics. To our knowledge ours is the first paper to address the problem with modern deep learning techniques and the first to model its heteroscedasticity.
The idea of neural heteroscedastic regression
was first proposed by \citet{nix1994estimating}, 
though they
do not share hidden layers between the two outputs, and are only concerned with Gaussian predictive distributions.
\citet{williams1996using} use a 
shared
hidden layer and consider the case of multivariate Gaussian 
distributions, for which they predict the full covariance matrix via its Cholesky factorization.
Heteroscedastic regression has a long history outside of neural networks. \citet{le2005heteroscedastic} address a formulation for Gaussian processes. 
Most related is
\citet{lakshminarayanan2016simple} which also revisits heteroscedastic neural regression,
also using a softplus activation to enforce non-negativity.
We show some successful modifications to the above work, such as the 
use of the Laplace distribution, 
but our more significant contribution is the application of the idea to clinical medical data.


\section{Discussion}
\label{sec:discussion}
Our results demonstrate both the efficacy of machine learning (over current approaches) and the heteroscedasticity of surgery duration data. 
In this section, we explore both results in greater detail. Specifically, we analyze the models 
to see \emph{which features} are most predictive
and examine the uncertainty estimates to see \emph{how they might be used} in decision theory to lower costs.

\subsection{Feature Importance}
\begin{figure*}[t]    
  		\centering
		\includegraphics[scale=0.4]{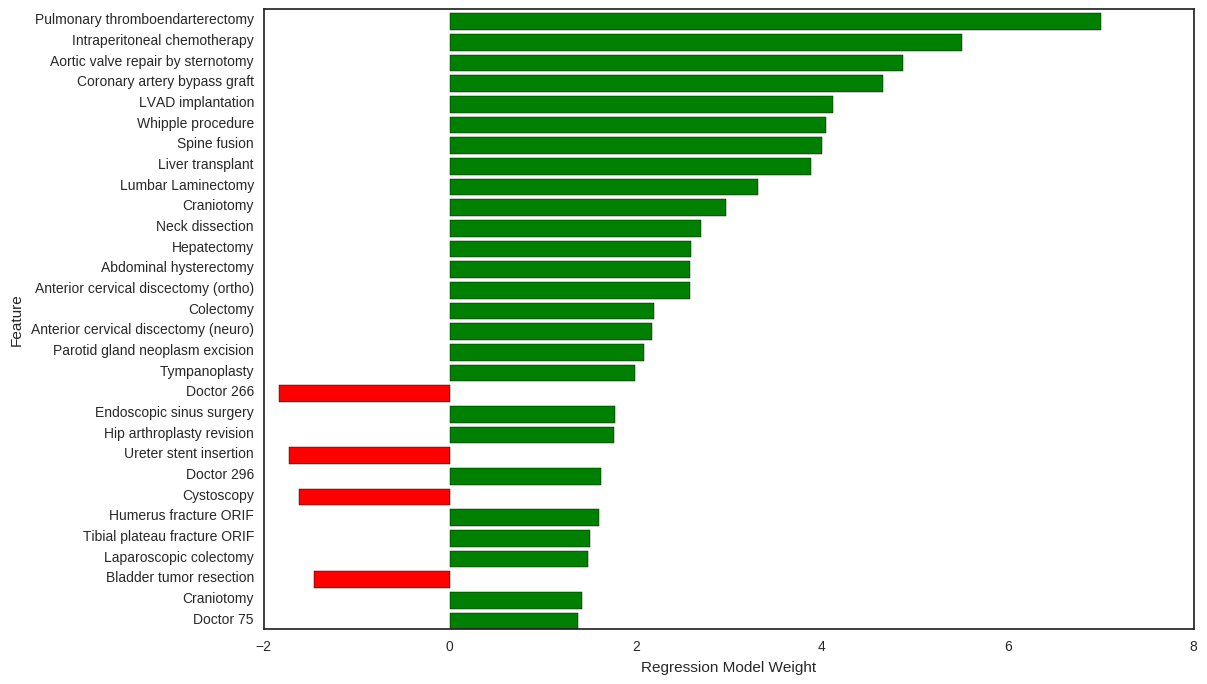}
		\caption{
Top 30 linear regression features sorted by coefficient magnitude.
        }
\label{fig:reg-weights}
\end{figure*}

First, we consider the importance of the various features. 
Perhaps the most common way to do this is to see which features corresponded to the largest weights in our linear model. These results are summarized in Figure \ref{fig:reg-weights}.
Not surprisingly, the top features 
are dominated by procedures. 
In particular pulmonary thromboendarterectomy 
receives the highest positive weight. 
This procedure involves a high risk of mortality,
a full cardiopulmonary bypass, hypothermia and full cardiac arrest. 
Interestingly, even after accounting for procedures,
two doctors receive high weight. 
One (Doctor 266) receives significant negative weight, indicating unusual efficiency and another (Doctor 296) appears to be unusually slow. 
For ethical reasons, we maintain the anonymity of both the doctors and their specialties. 

\begin{figure*}[t]    
  		\centering
		\includegraphics[scale=0.55]{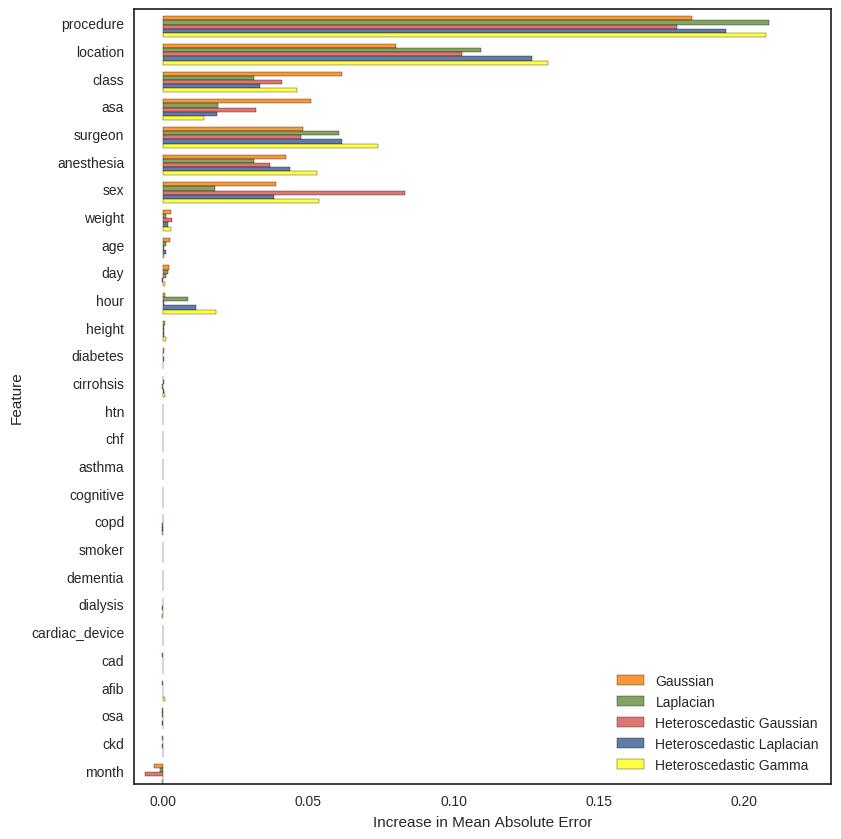}
		\caption{
Ablation analysis of feature importance for neural models.
        }
\label{fig:ablation}
\end{figure*}

For neural network models, 
we evaluate the importance of each feature group 
by performing an ablation analysis (Figure \ref{fig:ablation}). 
As a group, procedure codes are again the most important features. 
However, location, patient class, surgeon, anesthesia, and patient sex all contribute significantly.
The hour of day appears to influence 
the performance of our models 
but the day of the week does not 
and the month appears to merely introduce noise, 
leading to a reduction in test set performance.
Interestingly, comorbidities also made little difference in performance. However, it is possible that these features only apply to a small subset of patients but are highly predictive for that subset.

\subsection{Economic Analysis}
\begin{figure*}[ht!]    
  		\centering
		\includegraphics[scale=.6]{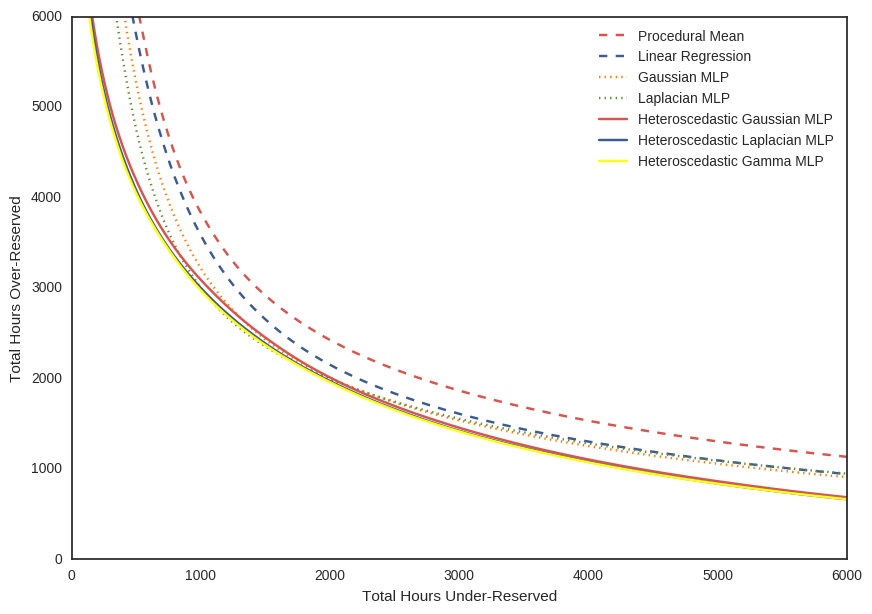}
		\caption{
Total over-booking and under-booking errors for different models as we adjust predicted values. Predictions were selected by considering different percentiles of the distribution predicted for each case. For homoscedastic models, this was equivalent to multiplying each prediction by a constant value.
        }
\label{fig:over-under-plot}
\end{figure*}

Our aim in predicting the variance of the error is to provide uncertainty information that could be used to make better scheduling decisions.
To compare the various approaches from an economic/decision-theoretic perspective, 
we might consider the plausible case where 
the cost to over-reserve the room by one minute (procedure finishes early) 
differs from the cost to under-reserve the room (procedure runs over). 
We demonstrate how the two quantities can be traded off in Figure \ref{fig:over-under-plot}.

For models that don't output variance, 
we consider scheduled durations
of the form $\hat{y}+k$ and $\hat{y} \cdot k$ where $k$ is a data-independent constant.
In either case, by modulating $k$, one books more or less aggressively. 
The multiplicative approach performed better, 
likely because long procedures have higher variance than short ones. This approach is equivalent to selecting a certain percentile of each predicted distribution given a constant sigma.

For heteroscedastic models we make the trade-off
by selecting a constant percentile 
of each predicted distribution. 
When the cost of over-reserving by one minute and under-reserving by one minute are equal, the problem reduces to minimizing absolute error. 
Around this point on the curve the homoscedastic Laplacian outperforms all other models. 
However, given cost sensitivity, 
the heteroscedastic Gamma strictly outperforms all other models.

\subsection{Future Work}
We are encouraged by the efficacy of simple machine learning methods both to predict the durations of surgeries and to estimate our uncertainty. 
We see several promising avenues for future work.
Most concretely, we are currently engaged in discussions 
with the medical institution whose data we used about introducing a trial in which surgeries would be scheduled according to decision theory based on our estimates. 
Regarding methodology,
we look forward to expanding this research in several directions.
First, we might extend the approach to modeling covariances and more complex interactions among multiple real-valued predictions. 
We might also consider problems like bounding box detection, requiring more complex neural architectures.


\bibliography{heteroscedastic_surgery}
\bibliographystyle{plainnat}


\end{document}